# Adaptive Decision Making at the Intersection for Autonomous Vehicles Based on Skill Discovery*

Xianqi He, Lin Yang, Chao Lu*, Zirui Li and Jianwei Gong, *Member, IEEE*

*Abstract*—In urban environments, the complex and uncertain intersection scenarios are challenging for autonomous driving. To ensure safety, it is crucial to develop an adaptive decision making system that can handle the interaction with other vehicles. Manually designed model-based methods are reliable in common scenarios. But in uncertain environments, they are not reliable, so learning-based methods are proposed, especially reinforcement learning (RL) methods. However, current RL methods need retraining when the scenarios change. In other words, current RL methods cannot reuse accumulated knowledge. They forget learned knowledge when new scenarios are given. To solve this problem, we propose a hierarchical framework that can autonomously accumulate and reuse knowledge. The proposed method combines the idea of motion primitives (MPs) with hierarchical reinforcement learning (HRL). It decomposes complex problems into multiple basic subtasks to reduce the difficulty. The proposed method and other baseline methods are tested in a challenging intersection scenario based on the CARLA simulator. The intersection scenario contains three different subtasks that can reflect the complexity and uncertainty of real traffic flow. After offline learning and testing, the proposed method is proved to have the best performance among all methods.

## I. INTRODUCTION

In urban environments, intersection scenarios are challenging for intelligent vehicles because of their uncertainty and complexity. As shown in Fig. 1, intelligent vehicles have to deal with various situations involving other vehicles, such as other vehicles changing lanes, other vehicles turning around, unprotected left turns with oncoming traffic, etc. To ensure safety, it is crucial to develop an adaptive decision making system that can handle the interaction with other vehicles.

Model-based methods have been applied to build decision making systems [1]. Time-to-collision (TTC) [2] is a safety indicator that is widely used [3]. With TTC, manually designed hierarchical state machines were adopted as the decision making mechanism for intersections in the DARPA Urban Challenge [4]. Generally, these methods are reliable and predictable. However, the model-based methods often

*This work was supported by the National Natural Science Foundation of China under Grants 61703041 and U19A2083, Ministry of Education (Project Grant No. 2017CX02005).

Xianqi He, Lin Yang, Chao Lu*, Jianwei Gong are with the School of Mechanical Engineering, Beijing Institute of Technology, Beijing 100081, China (email: 3120200328@bit.edu.cn; yanglin@bit.edu.cn; chaolu@bit.edu.cn; gongjianwei@bit.edu.cn)

Zirui Li is with the Department of Transport and Planning, Delft University of Technology, Delft 2628 CD, The Netherlands, and also with the with the School of Mechanical Engineering, Beijing Institute of Technology, Beijing 100081, China. (email: z.li@bit.edu.cn)

(corresponding author: Chao Lu)

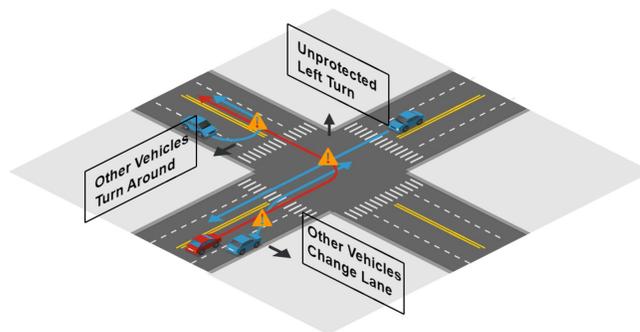

Figure 1. The intersection scenario

require manually designed driving strategies. Yet the number of new scenarios is large while the labor available for method designing is limited.

These limitations might be avoided with learning based methods. Imitation learning (IL) [5-9] learns driving policies from expert driving data, and it does not require manually designed strategies. However, the policies learned by IL are not satisfying when faced with situations outside of the training data. Besides, IL needs a large amount of expert data that are expensive and difficult to collect. Combing with neural networks, reinforcement learning (RL) becomes popular [10-12]. Unlike model-based methods and IL [13, 14], RL can adapt to changing environments by trial and error. In addition, some researchers also use RL to build human-like behavior learning systems that learn from human data [15, 16]. It is noteworthy that the RL can learn policies autonomously without labeled data, but they cannot be practically applied to complex and uncertain intersections because of the following reasons: 1) The learned knowledge is not transferable. As different scenarios have different characteristics, an RL algorithm trained for a specific scenario requires retraining to perform appropriately in other scenarios. 2) The RL methods are not interpretable. They usually use a single neural network to fit the problem, and the inner part of the neural network is a black-box structure with poor interpretability. As a result, it is difficult to analyze the neural networks and generalize them to new environments.

Hierarchical reinforcement learning (HRL) is introduced to solve these problems [17, 18]. RL methods with flat policy tend to be a smooth mapping, so it is difficult for them to solve complex problems with distinct subtasks. HRL introduces hierarchical structures and temporal abstraction into learning, which makes it possible to learn policies separately for distinct subtasks [19]. Moreover, some researchers combine the idea of motion primitives (MPs) with HRL [20]. MPs-based methods decompose a complex problem into multiple easy subtasks and arrange corresponding MPs for each subtask.

These MPs-based HRL methods have strong interpretability and scalability. However, the existing MPs-based HRL methods also suffer from the drawback that the MPs are manually defined, so they do not work well in unexpected situations.

In this paper, we propose an MPs-based HRL framework that enables autonomous learning of MPs. It is an adaptive decision making system that can reuse accumulated interpretable MPs, where MPs are learned autonomously based on skill discovery methods. With these features, the proposed method can deal with complex traffic scenarios with high uncertainty.

The remainder of the paper is organized as follows. Section II describes the proposed MPs-based HRL framework. In Section III, an intersection scenario is designed in CARLA and the details of the experiments are depicted. In section IV, the results of the proposed method and other baseline methods are discussed. Section V concludes the work.

## II. HIERARCHICAL REINFORCEMENT LEARNING WITH MOTION PRIMITIVES

### A. Hierarchical Decision Making System

As shown in Fig. 2, this paper utilizes a hierarchical architecture that divides the system into an upper decision making layer and a lower execution layer:

*1) Decision Making Layer:* The upper decision making layer uses the MPs automatically learned in the lower execution layer to build an extensible and interpretable MPs library. The MPs manager will observe the environment and determine whether the current subtask has been learned. If the subtask has already been learned, it selects the corresponding MP from the MPs library and reuses it. If the subtask is new, it autonomously creates a new MP and adds it to the MPs library after the training of MP is complete.

*2) Execution Layer:* The lower execution layer uses a skill discovery method to autonomously learn MPs, where every MP contains initiation set, termination set, and intra policy. The initiation set and termination set determine when the MP starts and when the MP stops, respectively. The intra policy is the strategy responsible for specific subtask assigned by the MP manager.

### B. Hierarchical Reinforcement Learning with MPs

In RL, a Markov decision process (MDP) consists of a set of states $\mathcal{S}$, a set of actions $\mathcal{A}$, a transition function $P:\mathcal{S}\times\mathcal{A}\times\mathcal{S}\to[0,1]$ and a reward function $r:\mathcal{S}\times\mathcal{A}\to\mathbb{R}$. We tend to maximize the cumulative reward, i.e., *expected return*, defined as $G_\pi(s)=\mathbb{E}_\pi\left[\Sigma_{t=0}^{\infty}r_{t+1}\mid s_0=s\right]$. In continuous tasks, the return can easily be infinite. Therefore, we use the discounted return $G_\pi(s)=\mathbb{E}_\pi\left[\Sigma_{t=0}^{\infty}\gamma^t r_{t+1}\mid s_0=s\right]$, where $\gamma\in[0,1)$ represents the *discount factor*, and $\pi$ represents the policy.

In details, the action at time step $t$ is denoted as $\mathbf{a}_t=[a_{t,\text{throttle}},a_{t,\text{brake}},a_{t,\text{steer}}]$ where $a_{t,\text{throttle}}$, $a_{t,\text{brake}}$ and $a_{t,\text{steer}}$ are the throttle, brake and steering angle of the host vehicle, respectively. The state $\mathbf{s}_t$ at time step $t$ is denoted as:

$$\begin{aligned}\mathbf{s}_t &= [\mathbf{s}_{t,\text{host}}\parallel\mathbf{s}_{t,\text{other}}]\\ \mathbf{s}_{t,\text{host}} &= [x_{t,\text{host}},y_{t,\text{host}},\theta_{t,\text{host}},v_{t,\text{host}}]\\ \mathbf{s}_{t,\text{other}} &= [x_{t,\text{other}},y_{t,\text{other}},\theta_{t,\text{other}},v_{t,\text{other}}]\end{aligned} \quad (1)$$

where $\mathbf{s}_{t,\text{host}}$ is the state of the host vehicle. The operator $\parallel$ stands for concatenation operation. $\mathbf{s}_{t,\text{other}}$ is the state of the vehicle closest to the host vehicle. $(x,y)$ is the position of the vehicles. $\theta$ and $v$ are the yaw angle and velocity of the vehicles, respectively.

As for MPs, each of them is a kinematically or dynamically feasible control sequence/trajectory that connects a pair of start/goal configurations [21]. We combine the idea of MPs with the options framework in HRL [19], which means every MP is an option. A Markovian option $\omega\in\Omega$ is a 3-tuple $(\mathcal{I}_\omega,\pi_\omega,\beta_\omega)$ in which $\mathcal{I}_\omega\subseteq\mathcal{S}$ is an initiation set, $\pi_\omega:\mathcal{S}\times\mathcal{A}\to[0,1]$ is an intra-option policy, and $\beta_\omega\subseteq\mathcal{S}$ is the termination set. In order to achieve autonomous learning of MPs, a skill discovery method called *deep skill chaining* is adopted [22] which will be explained later.

### C. Execution Layer Policy

As discussed above, We build the MP framework based on options. For the policy $\pi$ of MP, we use the model-free reinforcement learning method deep deterministic policy gradient (DDPG) to estimate the Q value $Q_\pi(s,a)$ [23]. The DDPG includes an online Q network and a policy network with parameters $\theta$ and $\phi$, as well as a target Q network and a policy network with parameters $\theta'$ and $\phi'$. The Q value is updated by using the following expression:

$$y = r + \gamma Q(s', \pi(s'|\phi')|\theta') \quad (2)$$

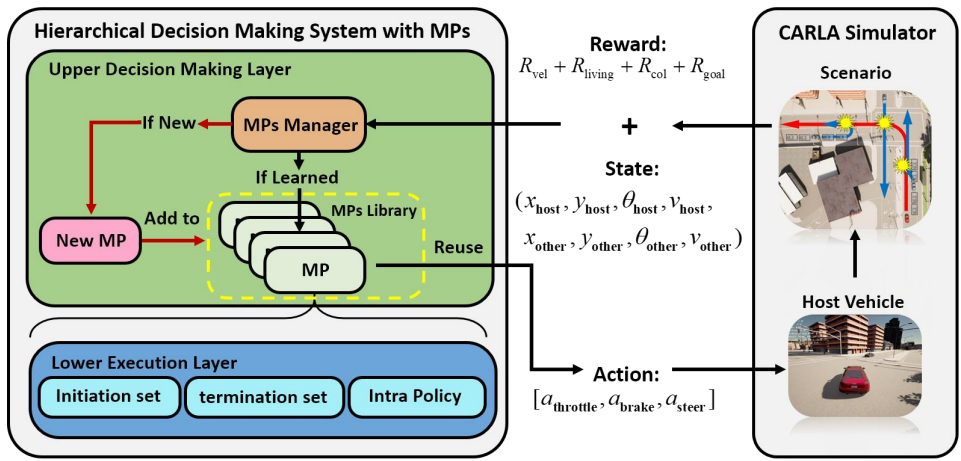

Figure 2. The hierarchical decision making system based on motion primitives

The critic is updated by minimizing the loss as:

$$L = \frac{1}{N}\sum_i (y_i - Q(s_i, a_i | \theta))^2 \quad (3)$$

The actor policy is updated as:

$$\nabla_{\theta\phi} J = \frac{1}{N}\sum_i \nabla_a Q(s,a|\theta)|_{s=s_i, a=\pi(s_i|\phi')} \nabla_{\theta\phi}\pi(s|\phi)|_{s_i} \quad (4)$$

The target networks under the "soft" update rule are updated as follows:

$$\begin{aligned} \theta' &\leftarrow \tau\theta + (1-\tau)\theta' \\ \phi' &\leftarrow \tau\phi + (1-\tau)\phi' \end{aligned} \quad (5)$$

with $\tau \ll 1$.

### D. Autonomously Learn MPs

In a high-dimensional continuous space, it is difficult to reach the endpoint. To ensure that the algorithm reaches it, a strategy named *deep skill chaining* for learning MPs from the endpoint to the start point is adopted. The core idea is that it is simpler to get to the end at the local neighborhood of the end [24]. Firstly, the MPs should learn the initial set with a certain size. Please note that the size of the set cannot be too large or too small. If the size is too small, it is difficult to trigger the MP. But if it is too large, the MP may be triggered in an inappropriate state. We train a one-class SVM classifier to autonomously learn it, and the random walk strategy is used to collect data: if the agent starts at state $s_t$, and reaches the termination set in $K$ timesteps, then we label it as a positive sample, otherwise, it is labeled as a negative sample. When $N$ samples have been collected, we use them to train the classifier to represent the initial set and the first MP is complete.

Then we learn the second MP. To ensure that the MPs can form an uninterrupted sequence, let the termination set of the second MP be equal to the initiation set of the first MP. Repeat and learn 3, 4, 5...n until the initial set of MP contains the starting point, i.e., finally complete the MP sequence from the starting point to the ending point.

### E. Decision Making Layer Policy

When the MP manager observes the environment, there may be several available MPs in state $s$. To solve it, HRL is used to select the best MP. We define a policy $\mu : S \times O_s \to [0,1]$, where $O$ represents the set of all options, $O_s$ represents the set of options available in state $s$, and $\mu(s,o)$ represents the probability of selecting $o$ as the current option in state $s$. The $Q$ function is mathematically expressed as:

$$Q^\mu(s,o) = \mathbb{E}_\pi\left[r_t + \gamma r_{t+1} + \gamma^2 r_{t+2} + \cdots | s_0 = s, o_0 = o\right] \quad (6)$$

The goal of the decision making layer is to maximize $Q^\mu(s,o)$ to find optimal solution. We use double deep Q-learning (DDQN) to evaluate the value of each MP and choose the most valuable one [25]. We use a deep neural network to represent the Q value as $Q(s,o|\theta)$ with parameters $\theta$, and represent the target Q as $Q(s,o|\theta')$ with parameters $\theta'$. The targets are mathematically defined as:

$$y = r + \gamma Q(s', \arg\max_{o'}(s',o'|\theta)|\theta') \quad (7)$$

We update the neural networks by minimizing the following function:

$$\min_\theta \sum (y - Q(s,o|\theta)) \quad (8)$$

Please note that a "soft" update rule, similar to DDPG, is also used.

## III. EXPERIMENTS

### A. Simulation Environment and Scenario

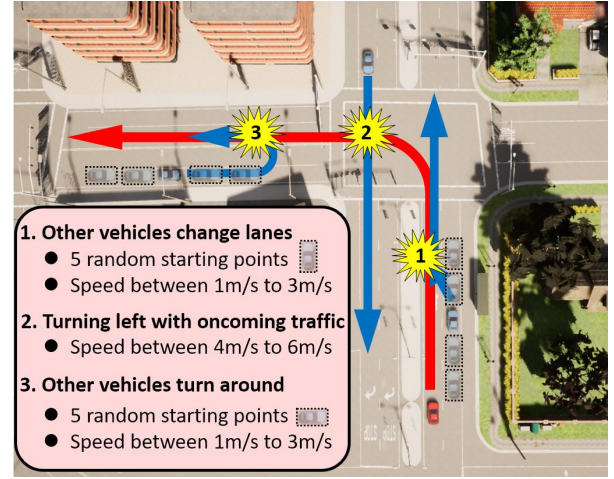

Figure 3. The experiment scenario built in CARLA

We test the proposed algorithm on the CARLA simulator, which provides a high-definition display, multiple sensors, and simulation with dynamics [26].

To test the proposed algorithm, we need to design a scenario that is interactive, uncertain, and complex. To fulfill these requirements, we designed an intersection scenario, as shown in Fig. 3. The host vehicle is denoted in red, and other vehicles are denoted in blue. There are three subtasks in this scenario: 1) Other vehicles change lanes. 2) turning left yielding to oncoming traffic. 3) Other vehicles turn around. In these subtasks, all other vehicles will have a 50% chance of moving and a 50% chance of stopping in place. If the other vehicles choose to move, they randomly choose the start point and speed, as shown in Fig. 3.

For the host vehicle, the algorithm controls longitudinal movement, i.e., accelerator and brake, and lateral movement is controlled by a pure pursuit controller. Other vehicles follow predetermined trajectories, where pure pursuit controllers control lateral movement and PID controllers control longitudinal movement.

### B. Offline Learning and Testing

Offline learning is necessary for MPs. In the beginning, the scenario is initialized without introducing other vehicles. We train all the MPs for 400 epochs so that they learn to drive smoothly at a goal speed of $v_{goal}$=5 m/s . Afterward, we train every MP separately, the training process also continues for 400 epochs for each subtask. Thus, offline learning has a total of 1600 epochs. To ensure fairness, all baseline methods will share the same offline learning process with the proposed method.

After the offline learning, the testing with all subtasks is performed. The length of testing is 1000 epochs, that is, the testing in the 1601~2600 epoch. There are all three subtasks in the scenario, each subtask has a half probability of appearing or not appearing. The host vehicle needs to flexibly adjust between the two strategies of avoiding other cars.

### C. Network Architectures

As shown in Table. I, all the networks are trained with Adam optimizer and soft update. The details of network architecture are as follows:

TABLE I. Hyperparameters of the decision making and execution layer

| Hyperparameter | Decision-making (DDQN) | Execution (DDPG) |
|---|---|---|
| Number of hidden layers | 2 | 2 (actor and critic) |
| Hidden size 1 | 256 | 400 |
| Hidden size 2 | 128 | 300 |
| Learning rate | 1e-4 | 1e-4 (actor) 1e-3 (critic) |
| Batch size | 64 | 64 |
| $\gamma$ | 0.99 | 0.99 |
| $\tau$ | 1e-3 | 0.01 |

*1) Decision Making Layer:* Based on the semi-Markov decision process (SMDP), a single DDQN network is used to control the policy over MPs. Each hidden layer is followed by a ReLU activation function. The final layer is slightly different as the number of nodes in this layer can be changed. With $n$ MPs, there are $n$ nodes in the final layer, each representing the Q-value of the corresponding action. When a new MP is created, we add a new node in the final layer.

*2) Execution Layer:* Based on MDP, a DDPG network is used to control the intra policy for every MP. With $n$ MPs, there are $n$ DDPG networks. In actor network, Each hidden layer is followed by ReLU activation and batch normalization. The final layer has 1 node corresponding to the size of the action, which is followed by a tanh activation function to limit the output range. In critic network, each hidden layer is followed by a ReLU activation function.

### D. Baseline Methods

We choose the following baseline methods:

*1) Flat Policy:* To compare with the proposed hierarchical method, DDPG is selected as the flat policy method. It shares the same architecture as the motion primitives in the hierarchical method.

*2) Tabular Method:* Q Learning is also selected to compare with the proposed method.

### E. Implementation Details

*1) Rewards Design:* After extensive testing, we found that the following reward function works well in practice:

$$R = R_{vel} + R_{living} + R_{col} + R_{goal} \tag{9}$$

$$R_{vel} = \begin{cases} 0.25v, & \text{if } v \leq 5 \\ 0.25(10-v), & \text{if } v > 5 \end{cases} \tag{10}$$

$$R_{living} = -0.5 \tag{11}$$

$$R_{col} = -100 \text{ if collision} \tag{12}$$

$$R_{goal} = 10 \tag{13}$$

where $R_{vel}$ encourages the host vehicle to keep moving forward with a target speed of 5 m/s . If the speed of the host vehicle exceeds the target speed, the reward will be gradually reduced, and if the speed is too high, penalize the host vehicle with a negative reward. $R_{living}$ encourages completing the task as quickly as possible. Without this restriction, the host vehicle may stop frequently. $R_{col}$ penalizes collision with other vehicles or buildings. In our case, we treat collision as a serious mistake and penalize it with a large negative reward. $R_{goal}$ reward the host vehicle for reaching the end.

*2) Exploration Strategies:* For the upper decision making layer and lower execution layer, we use Epsilon-Greedy as the exploration strategy. Epsilon-Greedy is a simple but effective method to balance exploration and exploitation by choosing between exploration and exploitation.

## IV. RESULTS

The experiment is divided into two parts: offline learning and testing. The testing scenario will contain all subtasks with additional randomness.

### A. Offline Learning

Fig. 4 shows the average return of MPs and baseline methods. Among the four stages, MPs have the best performance. Each motion primitives focuses on its own task, so the learning curve converges faster. DDPG performs similarly to MPs in most subtasks, except for the turning around subtask. This might be because it is the most difficult, and DDPG is powerless after learning three subtasks. As for Q Learning, it barely learns knowledge. The return of Q Learning almost keeps unchanged.

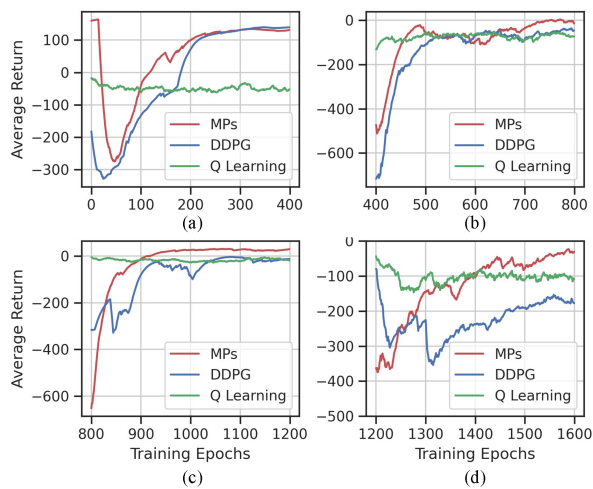

Figure 4. The Learning curves of different methods in offline learning. (a) The host vehicle drives at constant speed. (b) The other vehicles change lanes. (c) Turning left with oncoming vehicles. (d) The other vehicles turning around. The curves are the average of 3 runs. Exponential moving average (EMA) is used for visual clarity, where the weight is 0.95. Each epoch contains a maximum of 1000 steps

### B. Testing with all subtasks

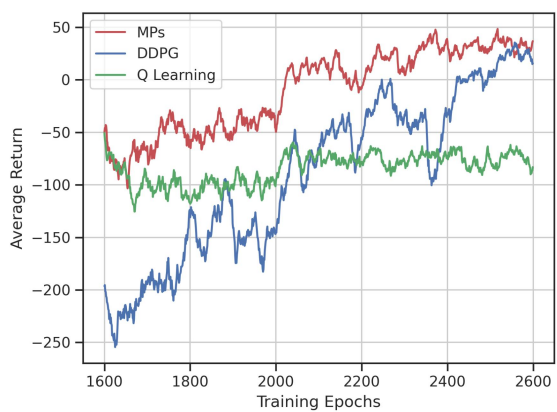

Figure 5. The Learning curves of different methods in testing. The curves are the average of 3 runs. Exponential moving average (EMA) is used for visual clarity, where the weight is 0.95. Each epoch contains a maximum of 1000 steps.

The test result is shown in Fig. 5. MPs perform best because they can reuse accumulated knowledge. When the scenario switches from offline learning to testing, MPs remember learned knowledge and keep good performance. Therefore, they have the advantage in the initial performance over the memoryless DDPG which needs 500 epochs retraining to reach the initial performance of MPs. Q Learning, like the offline learning stage, barely learns any knowledge.

After the testing is finished, we then collect the success rates of each subtask for 200 epochs. As shown in Table. II, the success rates are recorded at every subtask and goal point. Please note that the success rates of Q Learning are deceptive. Q Learning cannot deal with continuous tasks and it randomly chooses between throttle and brake. Therefore, the speed is always maintained at a low level and fluctuates greatly, as shown in Fig. 6(a). As a result, this "lazy" strategy makes the host vehicle coincidentally avoids collisions, as shown in Fig. 6(b) and Fig. 6(c). As for MPs and DDPG, MPs reuse the accumulated knowledge and perform much better.

TABLE II. THE SUCCESS RATES OF EACH SUBTASK IN TESTING

| Method | MPs | DDPG | Q Learning* |
|---|---|---|---|
| Other vehicles change lane | 71% | 50% | 58% |
| Unprotected left turn | 68% | 49% | 55% |
| Other vehicles turn around | 40% | 28% | 27% |
| Goal | 40% | 28% | 27% |

*Please note that the success rates of Q Learning are deceptive.

The vehicle kinematic parameters are as shown in Fig. 6. In the first subtask, both MPs and DDPG follow the other vehicles, and their performance is similar. In the second subtask, MPs choose to pass as quickly as possible under the premise of ensuring safety, which brings a smooth heading curve as shown in Fig. 6(b). But for DDPG, it chooses to slow down and let other vehicles pass first. In the third subtask, MPs observe and make decisions faster. Besides, the speed curve is also smoother as shown in Fig. 6(a). Q Learning keeps creeping slowly through all subtasks.

## V. CONCLUSION

In this paper, a hierarchical framework that can reuse accumulated knowledge is proposed. The proposed method combines the idea of MPs with HRL. This feature solves the problem that RL needs retraining after the scenario changes. The proposed method and the baseline methods are tested in a challenging intersection scenario. After offline learning and testing, the proposed method is proved to have the best performance. However, there are also deficiencies in this work. Although the proposed method performs best, the success rate of reaching the goal is still relatively low. This is due to the randomness of reinforcement learning. In the future, we will explore various methods to make the accumulated MPs more stable.

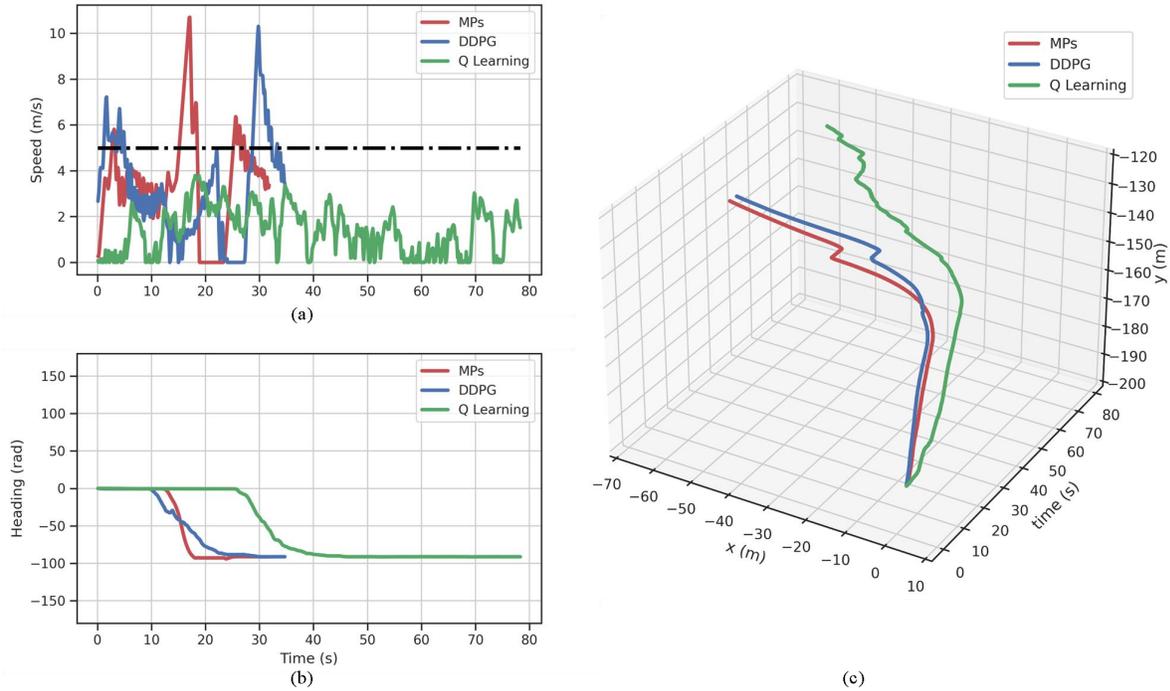

Figure 6. Kinematic parameters of the host vehicle. (a) The speed curves indicate the absolute velocity in the world coordinate system of the host vehicle. (b) The heading curves indicate the yaw angle in the world coordinate system of the host vehicle. (c) The spatio-temporal trajectories in the world coordinate system of the host vehicle.